\documentclass[preprint,12pt]{elsarticle}

\usepackage{amssymb}
\usepackage{graphicx}    % Required to insert images
\usepackage{subcaption}  % Required for subcaptions
\usepackage{algorithm}
\usepackage{algorithmic}
\usepackage{amsmath}
\usepackage{multirow}
\begin{document}

\begin{frontmatter}



\title{Ellipse Meets Bit-Planes: A Novel Approach to RNFL based Glaucoma Detection Using Advanced Image Processing and Deep Learning}


\author[inst1]{Snigdha Paul}

\affiliation[inst1]{organization={Electronics and Communication Engineering},%Department and Organization
            addressline={Heritage Institute of Technology}, 
            city={Kolkata},
            postcode={700107}, 
            state={West Bengal},
            country={India}}

\author[inst1]{Sambit Mallick}
\author[inst1]{Anindya Sen}

\begin{abstract}
This work proposes an integrated pipeline for automatic glaucoma detection method from easily available colour fundas images based on an adaptive algorithm for ellipse-based polar transformation, to enhance the analysis of the Retinal Nerve Fiber Layer (RNFL) as the primary biomarker for observing glaucomatous changes, regardless of optic disc and macula position. Utilizing this transformation, we introduce two distinct frameworks tailored to different operational needs. The first framework, a deep learning-inspired feature fusion approach, achieves a 99.3\% detection rate, ideal for settings where high precision is essential, despite higher computational demands. The second framework employs a novel image-processing algorithm based on bit-plane slicing, offering 92.31\% accuracy and optimized for environments requiring rapid inference with minimal resource consumption. Both frameworks provide scalable and cost-effective solutions for early glaucoma detection. This study highlights the potential of RNFL-based diagnostic tools in addressing the global challenge of glaucoma, particularly in underserved regions.
\end{abstract}

\begin{keyword}
Glaucoma \sep Fundas image \sep Retinal Nerve Fibre Layer \sep Image processing \sep Feature extraction \sep Deep learning
\end{keyword}

\end{frontmatter}


\section{Introduction}
\label{sec:introduction}

Glaucoma is one of the leading cause of blindness in world \cite{Quigley2006, Resnikoff2004}. Both open angle and angle closure glaucoma is asymptomatic in its early phase, making glaucoma identification and treatment difficult, thus most cases goes undetected and left undiagnosed until the advanced late stages. Early detection of glaucoma is crucial as it allows timely intervention to prevent irreversible vision loss and blindness. Morphological Retinal Nerve Fiber Layer(RNFL) defects is an indication for early stage of glaucoma \cite{Hoyt1973, Lee2004, Tatham2015}.

The RNFL is a critical structure located in the inner retina, consisting of axonal bundles from retinal ganglion cells. It extends from the retina to the optic nerve head(ONH), where it converges into the optic nerve. In glaucoma, RGC loss leads to atrophy of the RNFL, resulting in a decrease in thickness. This thinning manifests as localized defects or diffuse defects, often referred to as RNFL defects (RNFLD), which are significant early indicators of glaucomatous damage. These defects are associated with morphological changes in the ONH and corresponding visual field abnormalities in the later stage \cite{jones-odeh2015relationship}.

The RNFL is visualized through advanced imaging modalities such as red-free fundus photography, scanning laser polarimetry, and optical coherence tomography (OCT). The RNFL thickness is heterogeneously distributed across the retina, with the inferior and superior regions exhibiting the greatest thickness, particularly along the arcuate bundles. Conversely, the nasal and temporal regions generally present with a thinner RNFL compared to the superior and inferior zones. Traditionally, the ``ISNT rule'' (Inferior $>$ Superior $>$ Nasal $>$ Temporal) has been employed to describe the normative RNFL thickness distribution. However, recent studies propose the ``IST rule'' (Inferior $>$ Superior $>$ Temporal), which omits the nasal region from the analysis, thereby marginally enhancing the specificity for glaucoma detection at the expense of sensitivity \cite{pradhan2016isnt}. It is crucial to note that defects often initially manifest in the superior and inferior arcuate regions due to their inherently higher RNFL thickness, rendering these areas pivotal for the early identification of glaucomatous changes.

Research has predominantly focused on the diagnostic efficacy of OCT in glaucoma patients exhibiting definitive localized RNFL defects visible on fundus imaging. However, it is noteworthy that only 20\% of glaucoma patients are present with localized RNFL defects \cite{jonas1994localized}. Clinical detection of RNFL defects via fundus photography becomes feasible after approximately a 50\% loss of the RNFL, as indicated by prior studies \cite{quigley1982quantitative}. Consequently, this work emphasizes diffuse RNFL thinning, characterized by a widespread reduction in RNFL thickness across a more extensive area. Unlike localized RNFL defects which manifest as distinct wedge-shaped dark regions, diffuse RNFL thinning is more subtle and poses a greater challenge for detection.

The analysis of colour fundas image is a preferred modality of glaucoma assessment as it is non-invasive and economical, and hence a potential method for large scale mass screening programs. In fundus images, the RNFL bundle appears as bright striations, with the characteristic arcuate pattern reflecting the distribution of nerve fibers. This ``bright-dark-bright'' (Superior-Temporal-Inferior) pattern is considered normal, where the dark region is located between the ONH and the macula \cite{thomas2011evaluation}. In glaucomatous eyes, this normal arcuate pattern is often violated, with RNFL defects emerging.

This paper proposes a novel pipeline for automatic glaucoma detection centered on an adaptive algorithm for Ellipse-based Polar Transformation (EPT) to effectively visualize RNFL bundles. Unlike standard polar transformations, which struggle to follow the RNFL trajectory, this method straightens RNFL bundles regardless of the optic disc and macula positions. This alignment enhances calculations and visualizations, providing a clearer representation of RNFL changes, especially for detecting glaucomatous defects in diffuse RNFL thinning. This technique represents an advancement of the method introduced in \cite{muramatsu2010detection}, extending its capabilities to provide improved alignment and robustness in varied anatomical configurations.

Following the enhanced representation of RNFL bundles through the adaptive Ellipse-based Polar Transformation (EPT), this paper introduces two distinct pathways for glaucoma detection centered on diffuse RNFL loss. The first path presents a deep learning-inspired feature fusion methodology that amalgamates two distinct feature classes for classification. The initial feature set, extracted from the annular peripapillary zone using deep learning techniques, captures the progression patterns characteristic of glaucoma. The second feature set is derived from the processed polar-transformed image, where the vertical alignment of RNFL trajectories enhances the representation of features indicative of diffuse RNFL thinning. Techniques such as edge detection and bit-plane slicing further refine these features. Although resource-intensive—due to the high computational demands of training complex neural networks and processing large volumes of data—this approach seeks to achieve high accuracy by integrating deep learning-based feature extraction with traditional image processing, thereby providing a comprehensive solution for glaucoma detection.

The second path emphasizes an advanced image processing algorithm designed for faster and more resource-efficient detection of diffuse RNFL thinning. It leverages the vertically aligned image, where improved alignment of RNFL bundles enhances feature representation. The main contribution of this paper lies in the mathematical formulation of an adaptive threshold and balance factor, which are dynamically adjusted based on the retinal fundas image characteristics to optimize classification. This approach minimizes computational complexity while ensuring precise differentiation between glaucomatous and healthy eyes, thereby achieving competitive accuracy with reduced resource demands, making it a scalable solution for glaucoma detection.

This research concentrates on the prediction of glaucoma predicated solely on retinal nerve fiber layer (RNFL) thinning, deliberately isolating its influence on glaucoma progression without incorporating factors such as the cup-to-disc ratio or the ISNT rule. By honing in on RNFL thinning, the study enhances early glaucoma detection by directly correlating RNFL alterations with glaucomatous damage. The two proposed frameworks, which amalgamate advanced image processing techniques with deep learning-inspired methodologies, provide complementary solutions for precise and scalable detection. This approach proposes a significant contribution for a major advance in clinical diagnosis of glaucoma and reduces dependency on conventional diagnostic factors.

The paper is organized as follows: Section \ref{sec:related works} presents a detailed review of the background research. Section \ref{sec:methods} explains the proposed method in detail. Section \ref{sec:results} discusses the results and provides an in-depth analysis. Finally, Section \ref{sec:conclusion} concludes the paper.

\section{Related Works}
\label{sec:related works}
The detection of glaucoma through the analysis of the RNFL has been extensively researched, leveraging various imaging modalities and computational techniques to evaluate the structural and functional integrity of the RNFL. Foundational work in RNFL analysis has underscored its critical role in both the detection and monitoring of glaucoma progression.

Early pioneering studies by Lundstrom and Eklundh revealed that consecutive changes in the intensity of RNFL patterns are reliable indicators of glaucoma progression, utilizing computerized densitometry for single-patient analysis \cite{lundstrom1980computer}. Similarly, Peli et al. advanced this field by developing an image analysis technique to measure RNFL striations from digitized black-and-white fundus photographs, enhancing the detection and monitoring of progressive, diffuse RNFL loss \cite{peli1989computer}. 

Additionally, Storp et al. highlighted the variability in RNFL thickness related to optic disc size, presenting a significant challenge in accurately interpreting RNFL measurements for glaucoma detection, particularly in individuals with atypical optic disc sizes \cite{storp2023evaluation}.

Advancements in feature extraction methods have significantly propelled RNFL analysis capabilities in glaucoma detection, employing various texture and spectral techniques. Yogesan et al. utilized the gray level run length matrix (GLRLM) texture estimator, achieving an accuracy of 80–90\% in identifying RNFL loss, particularly for focal wedge-shaped defects surrounded by healthy nerve fibers \cite{yogesan1998texture}. However, this method faced limitations in detecting diffuse RNFL loss. 

Tuulonen et al. applied an information-theoretical approach using Kullback Information Distance (KID) to identify microtexture changes in RNFL digital images, providing compelling evidence of structural alterations associated with glaucoma \cite{tuulonen2000digital}.

In another significant contribution, Kolar and Jan combined fractal dimensions (FDs) and power spectral characteristics for glaucoma detection, demonstrating computational efficiency and effective differentiation between glaucomatous and healthy RNFL structures \cite{kolar2008detection}. Muramatsu et al. proposed an automated method utilizing Gabor filters to enhance RNFL patterns, showing strong performance for detecting focal RNFL losses \cite{muramatsu2010detection}, though with limited effectiveness in cases of diffuse RNFL damage. Similarly, Odstrcilik et al. introduced a texture analysis approach based on Markov random fields, enabling reliable classification of healthy RNFL areas and regions with focal RNFL loss \cite{odstrcilik2012analysis}.

Further refining RNFL analysis, Acharya et al. introduced higher-order spectra and co-occurrence matrices for RNFL texture analysis, achieving over 91\% specificity in detecting glaucomatous eyes \cite{acharya2011automated}. Similarly, Panda et al. proposed a novel method for RNFL defect detection and angular width quantification, incorporating blood vessel inpainting, CLAHE-based contrast enhancement, and advanced texture and intensity features, offering robust angular measurements for improved detection \cite{PANDA201856}. 

Despite these advancements, challenges in accurately detecting diffuse RNFL losses persist, emphasizing the need for enhanced methodologies. While previous studies have made significant contributions to glaucoma detection, many have not utilized advanced machine learning and deep learning techniques. In contrast, this research incorporates these methodologies, enhancing detection accuracy and robustness.

Polar transformation techniques have also been employed to improve RNFL defect detection. Oh et al. utilized a standard polar transformation in conjunction with edge detection to identify local RNFL defects \cite{Oh2015AutomaticCD}. However, their approach relied on a conventional polar transformation that did not align with the natural trajectory of the RNFL, potentially limiting its ability to capture finer structural details.

Considerable research has also been directed towards the use of OCT for automated RNFL-related glaucoma detection. Studies by Chan et al. \cite{CHAN2019103483}, Leung \cite{Leung2022RetinalNT}, and Chen et al. \cite{Chen02012025} leveraged OCT imaging to automate glaucoma detection, highlighting the effectiveness of this modality in analyzing RNFL abnormalities. While OCT is effective for glaucoma detection, this approach utilizes fundus photography, which is more cost-effective and widely accessible, especially in regions where OCT devices are limited.

Most deep learning approaches in RNFL analysis from fundas photography have also predominantly relied on corresponding OCT images for training and prediction. Manivannan et al. focused on RNFL visibility classification on fundas images, contributing to more targeted glaucoma detection analysis \cite{article}. Hemelings et al. demonstrated that deep learning models could successfully detect glaucoma from fundus image regions beyond the ONH, underscoring the potential of deep learning techniques in glaucoma diagnosis \cite{Hemelings2021}.

Xu et al. proposed a deep learning framework to predict RNFL thickness around the optic disc using OCT data as training targets \cite{XU2024e33813}. While their model incorporated patch-based training, the reported sensitivity and specificity were below 80\%, indicating opportunities for further optimization. Yang et al. designed convolutional neural network models to predict RNFL thickness both globally and regionally illustrating the utility of OCT data along with fundas image in deep learning-based RNFL analysis \cite{Yang2022}.

Prananda et al. demonstrated the application of various deep learning models on the ORIGA dataset, achieving up to 92\% accuracy by using cropped fundus images after optic disc and blood vessel removal \cite{Prananda2023}. Similarly, Ding et al. proposed a position-guided deep learning method for RNFL defect detection, incorporating both physiological positioning and global dependencies to achieve more precise and reliable results \cite{10.1007/978-3-030-59722-1_72}. These studies highlight the versatility and efficacy of deep learning models in addressing RNFL-related challenges, while emphasizing the potential for further advancements in model performance, generalization, and clinical applicability.

While the aforementioned studies have made notable strides in advancing the understanding and detection of RNFL-related glaucoma, gaps remain in several areas that our research aims to address. For instance, some approaches, like the one by Oh et al. \cite{Oh2015AutomaticCD}, employed a standard polar transformation that does not fully align with the natural trajectory of the RNFL, limiting its ability to capture finer structural details necessary for accurate detection. Our research seeks to bridge this gap by introducing a more tailored ellipse-based polar transformation technique that aligns with the RNFL's natural trajectory. 

Furthermore, while many deep learning studies have utilized OCT images as training targets, we propose both deep learning-inspired techniques and image processing algorithms without relying on OCT data as the primary target, offering a novel approach to RNFL-based glaucoma detection. These advancements underscore the potential for a more refined and comprehensive solution to the challenges associated with RNFL analysis and glaucoma detection.

\section{Materials and Methods}
\label{sec:methods}

\subsection{Dataset}
\label{subsec:dataset}
The ``FIVES'' dataset constitutes a comprehensive repository of fundus images, systematically categorized into four classes: normal, glaucoma, diabetic retinopathy, and age-related macular degeneration. As documented in the research paper \cite{jin2022fives}, all these images of four different classes were acquired from the Ophthalmology Centre at the Second Affiliated Hospital of Zhejiang University (SHAZU), involving 573 patients ranging from 4 to 83 years of age.

\begin{table}[H]
\centering
\caption{Summary of Selected Classes from the FIVES Dataset}
\begin{tabular}{|c|c|p{6cm}|}
\hline
\textbf{Class} & \textbf{Number of Images} & \textbf{Description} \\
\hline
Normal & 200 & Fundus images from individuals with no diagnosed eye disease \\
\hline
Glaucoma & 200 & Fundus images exhibiting characteristic signs of glaucoma \\
\hline
\end{tabular}
\label{tab:fives_subset}
\end{table}

Captured using TRC-NW8 fundus cameras with a 50° field of view (Topcon Medical Systems, Tokyo, Japan) and centered at the macula, the images are provided at a resolution of 2048 × 2048 pixels. Additionally, the dataset includes expertly annotated ground truth vessel segmentation masks, which serve as a critical asset for vascular analysis.

To enhance the study’s robustness, approximately 5\% of the dataset comprises low-quality images, deliberately included to simulate real-world clinical scenarios where variability in image quality is prevalent. For the purposes of this research, only the Glaucoma and Normal categories were analyzed, allowing for a focused examination of the diagnostic intricacies associated with glaucoma detection.

\subsection{Hardware and Software}
\label{subsec:hardware and software}

We utilized Google Colab, which provides free access to NVIDIA GPUs. This resource enables rapid training of deep learning models and significantly enhances computational performance for large datasets. In addition, we also used a standard laptop CPU for lighter processing tasks and preliminary experiments. Python was employed as the primary programming language, supported by powerful libraries. OpenCV facilitated high-level image processing and feature extraction, while PyTorch served as the framework for implementing and training deep learning models for tasks such as object detection and classification.

\subsection{Preprocessing}
\label{subsec:preprocessing}

\begin{figure}[H]
    \centering
    % First image
    \begin{subfigure}[b]{0.32\textwidth}
        \centering
        \includegraphics[width=\textwidth]{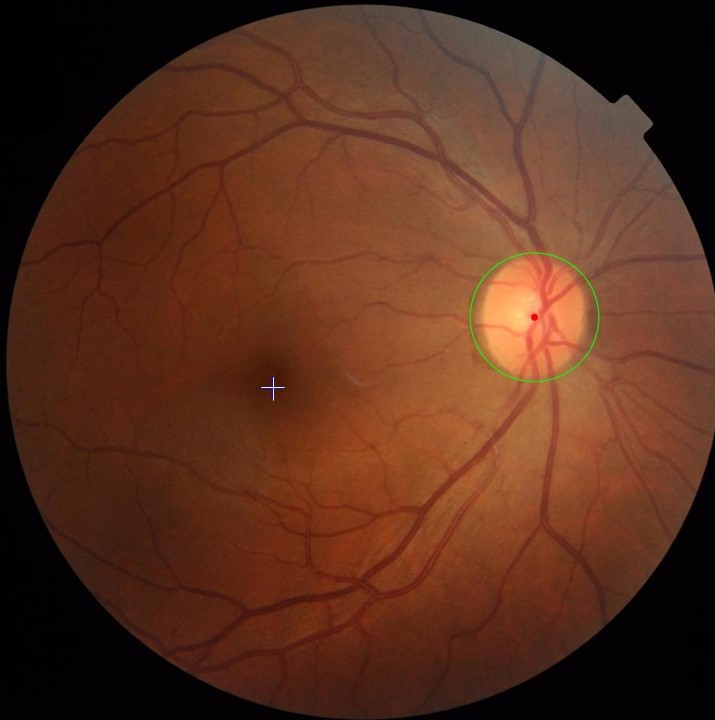} % Replace with your first image file
        \subcaption{}
        \label{fig:image1}
    \end{subfigure}
    \hfill
    % Second image
    \begin{subfigure}[b]{0.32\textwidth}
        \centering
        \includegraphics[width=\textwidth]{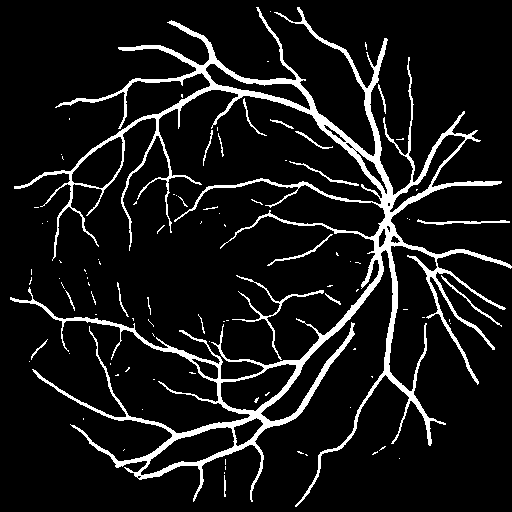} % Replace with your second image file
        \subcaption{}
        \label{fig:image2}
    \end{subfigure}
    \hfill
    % Third image
    \begin{subfigure}[b]{0.32\textwidth}
        \centering
        \includegraphics[width=\textwidth]{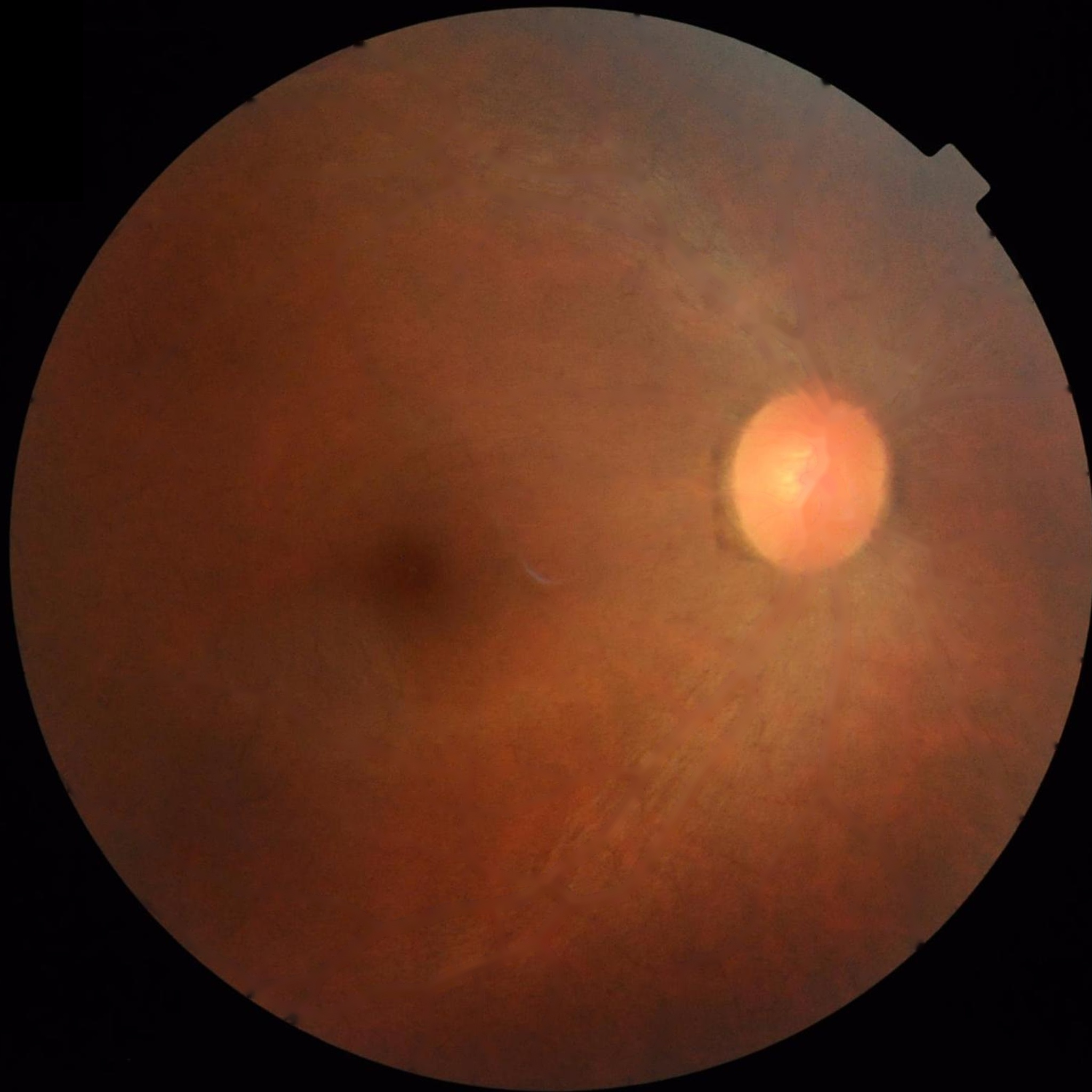} % Replace with your third image file
        \subcaption{}
        \label{fig:image3}
    \end{subfigure}
    
    \caption{
        (a) The red point indicates the optic disc center, the green circle represents the approximate boundary of the optic disc, and the white cross sign marks the macula point. (b) Vessel segmented output through U-Net with EfficientNetB7 backbone. (c) Inpainted image after vessel removal.
    }
    \label{fig:fundus_preprocessing}
\end{figure}

During the preprocessing of fundus images, the YOLOv8 object detection model was used to detect both the optic disc and the macula as shown in Figure \ref{fig:fundus_preprocessing}. A subset with 107 annotated optic disc samples and 176 annotated macula samples was taken to enrich the training data. The model was then trained on this augmented dataset to improve detection accuracy. 

The center of the predicted bounding box for the optic disc was designated as the optic disc center, with its radius inferred from the bounding box dimensions. Similarly, the center of the predicted bounding box was used as the macula's reference point, ensuring precise localization of these anatomical features.

To segment the retinal blood vessels, a U-Net architecture with an EfficientNetB7 encoder backbone was employed. The model was designed for pixel-wise segmentation, with EfficientNetB7 extracting features and incorporating skip connections from key layers. The decoder progressively upsampled the feature maps through the convolutional and activation layers, complemented by batch normalization. 

The final output was a binary mask representing the blood vessels, which was subsequently used as a mask for inpainting the fundus image using the fast marching method with a neighborhood radius of 10 pixels\cite{Telea2004}. This approach ensured precise segmentation of the vessels and restoration of regions obscured by the vasculature.

Following these steps, the green channel of the fundus images was extracted exclusively, as it provides optimal contrast for analyzing the structures of the retina. In addition, the peripapillary region, here referred to as the peripapillary retinal zone, was defined and extracted. The Peripapillary Retinal Zone lies outside the optic disc and extends within 7/10th of the distance between the optic disc center and the macula \cite{10.1007/978-3-030-59722-1_72}. This step ensured that the analysis focused on the region of interest most relevant for evaluating retinal health.

\subsection{Ellipse based Polar Transformation}

This approach introduces an adaptive ellipse-based polar transformation, which utilizes a simple equation to more accurately align with the RNFL trajectory within the peripapillary region \cite{muramatsu2010detection, Hemelings2021}, covering the temporal, supratemporal, and infratemporal bundle regions.

\begin{algorithm}[H]
\caption{Ellipse-Based Polar Transformation Algorithm}
\label{EPT}
\begin{algorithmic}[1]
\STATE \textbf{Input:} Optic disc center $(x_c, y_c)$, macula center $(x_m, y_m)$, retina center $(x_r, y_r)$, optic disc radius $r_{\text{optic}}$, and image $I(x,y)$
\STATE \textbf{Output:} Polar-transformed image $I_p(r,\theta)$
\STATE Compute the distance $d$ and angle $\theta_c$ between the optic disc and macula centers: $d = \sqrt{(x_c - x_m)^2 + (y_c - y_m)^2}$,\hspace{1em} $\theta_c = \tan^{-1}\left(\frac{y_m - y_c}{x_m - x_c}\right)$
\STATE Compute the distance $r$ from the optic disc center to the retina center: $r = \sqrt{(x_c - x_r)^2 + (y_c - y_r)^2}$
\STATE Compute the major axis $a$ and minor axis $b$ of the ellipse: 
\STATE \hspace{2em} $a = d + \lambda_1 \cdot \frac{C_{\text{max}}}{r} \cdot \frac{|\theta|}{90} + \lambda_2 \cdot \frac{1}{1 + \exp(-\gamma \cdot (|\theta| - \text{inflection\_angle}))} \cdot |\theta|$
\STATE \hspace{2em} $b = a \cdot \frac{|\theta|}{90}$
\STATE Use binary search to find the inflection angle $k$:
\STATE \hspace{1em} Initialize: $k \in [-30^\circ, 0^\circ]$, left=$-30^\circ$, right=$0^\circ$  and precision $\epsilon$=0.1.
\STATE \hspace{1em} While $|\text{right} - \text{left}| > \epsilon$:
\STATE \hspace{2em} Compute $a(k)$, $b(k)$, and the intersection distance $d(k)$ for angle $k$.
\STATE \hspace{3em} $d(k) = \left| \left|\text{angle of intersection} - \theta_c \right| - 30^\circ \right|$
\STATE \hspace{2em} If $d(k_{\text{mid}}) < d(k_{\text{left}})$, update: $\text{right} = k_{\text{mid}}$.
\STATE \hspace{2em} Else if $d(k_{\text{mid}}) < d(k_{\text{right}})$, update: $\text{left} = k_{\text{mid}}$.
\STATE \hspace{2em} Else, break.
\STATE \hspace{1em} Return optimal $k$ where $d(k)$ is minimized.
\FOR{each angle $\theta$ from $-90^\circ$ to $+90^\circ$ in steps of 0.1°}
    \STATE Compute the ellipse center $(x_e, y_e)$:
    \STATE \hspace{1em} $x_e = x_c + a \cdot \cos(\theta + \theta_c)$, \hspace{1em} $y_e = y_c + a \cdot \sin(\theta + \theta_c)$
    \STATE Determine half of the ellipse to draw based on $\theta + \theta_c$:
    \STATE \hspace{1em} If $x_c < x_m$: Draw lower half for $\theta \leq 0$, upper half for $\theta > 0$
    \STATE \hspace{1em} If $x_c > x_m$: Draw upper half for $\theta \leq 0$, lower half for $\theta > 0$
\ENDFOR
\STATE Define papillomacular region: $r_{\text{optic}} \leq r_p \leq 0.7 \cdot r$
\STATE Transform to polar coordinates: $r = \sqrt{\text{dx}^2 + \text{dy}^2}$, $\theta = \text{angle}$
\STATE Map pixel intensities: $I_p(r, \theta) = I(x, y)$
\STATE \textbf{Return:} Polar-transformed image $I_p$
\end{algorithmic}
\end{algorithm}

Unlike standard polar transformations or normal elliptical models, which do not follow the RNFL trajectory accurately, this method adapts to the transformation in the retinal color fundus images to fit the complex anatomy of the optic disc. The Algorithm \ref{EPT} outlines this adaptive transformation.

\begin{figure}[H]
    \centering
    \includegraphics[width=0.4\textwidth]{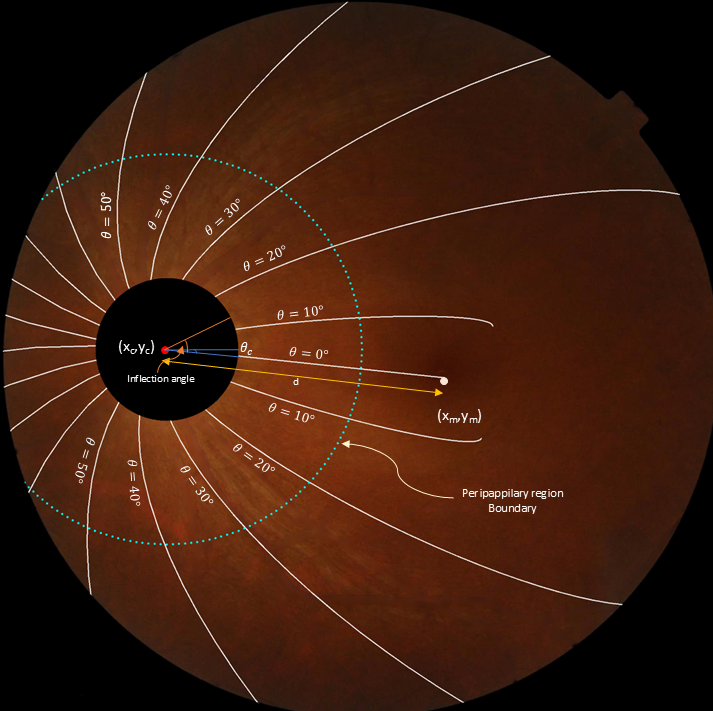} % Replace with your image file
    \caption{Parameters used in the ellipse-based transformation for aligning the RNFL trajectory within the peripapillary region.}
    \label{fig:fundus_marking}
\end{figure}

The ellipse generated at (in steps 4 -7 of Algorithm \ref{EPT}) varying angles of $\theta$ is defined by its major axis $a$ and minor axis $b$. The lengths of these axes are intricately dependent on the position of the optic disc center, macula fovea center, and the size of the optic disc. These dependencies are mathematically characterized as Equation \ref{eq:major_axis} and \ref{eq:minor_axis}:

\begin{equation}
a = d + \lambda_1 \cdot \frac{C_{\text{max}}}{r} \cdot \frac{|\theta|}{90} + \lambda_2 \cdot \frac{1}{1 + \exp(-\gamma \cdot (|\theta| - \text{inflection\_angle}))} \cdot |\theta|
\label{eq:major_axis}
\end{equation}

\begin{equation}
b = a \cdot \frac{|\theta|}{90}
\label{eq:minor_axis}
\end{equation}

where \( d \) represents the distance between the optic disc center and the macula center, and \( C_{\text{max}} \) denotes the maximum radius 1024, r is distance between the optic disc center to the retina center, $\theta$ varies from $-90^\circ$ to $90^\circ$ as shown in Figure \ref{fig:fundus_marking}. $\lambda_1$, $\lambda_2$ and $\gamma$ are scaling factors that are experimentally evaluated to the value 125, 15, 0.6, respectively, for this dataset. The parameters $\lambda_1$, $\lambda_2$ and $\gamma$ were optimized through a grid search over a range of plausible values. A validation set of 50 images was used to visually assess the generated ellipse fit to the RNFL arc. The selected values provided the best trade-off such that the summation of accuracy is maximised. Further fine-tuning on larger datasets is expected to enhance model robustness.

\begin{figure}[H]
    \centering
    % First row, first image
    \begin{subfigure}[b]{0.32\textwidth}
        \centering
        \includegraphics[width=\textwidth]{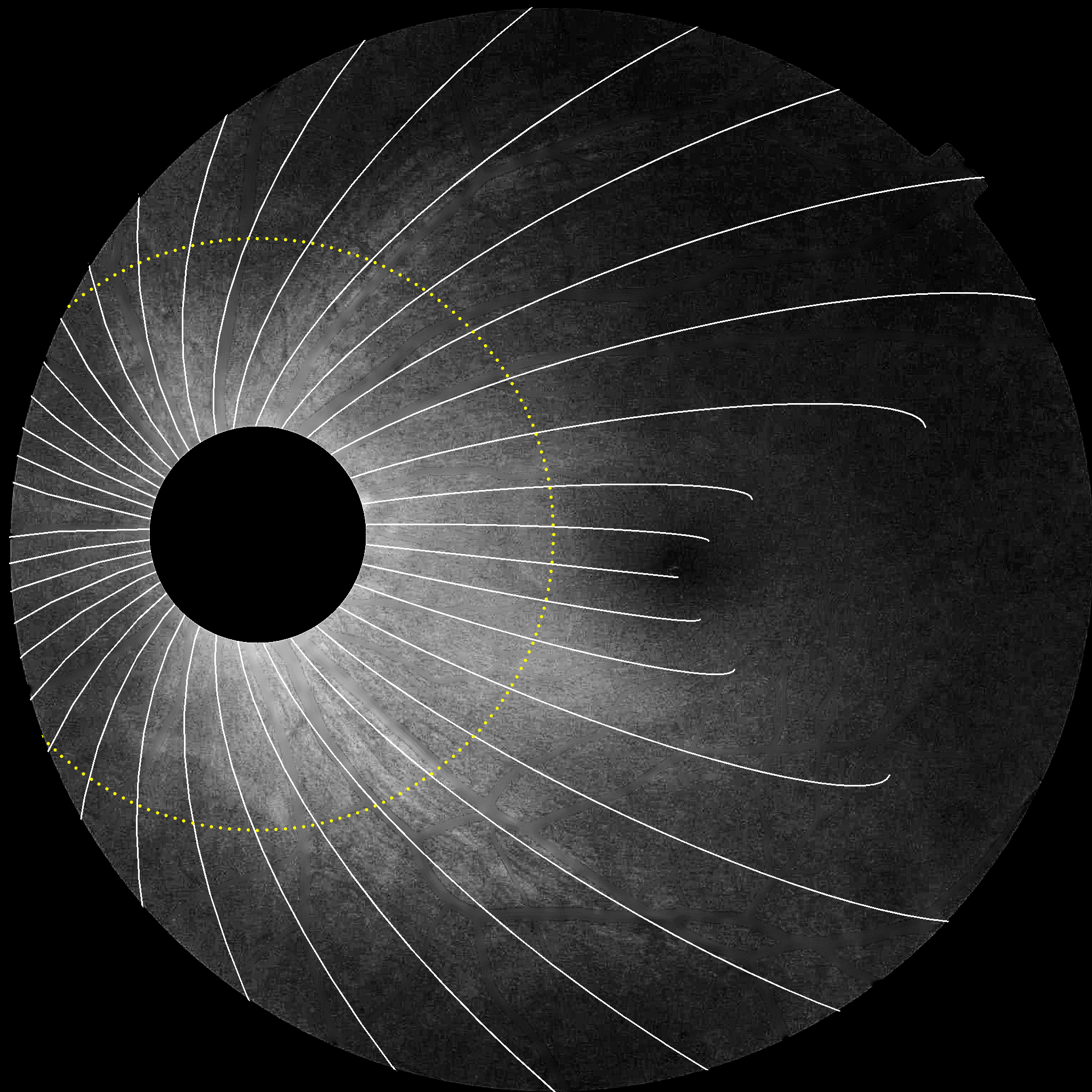} % Replace with your first image file
        \subcaption{}
        \label{fig:image4}
    \end{subfigure}
    \hfill
    % First row, second image
    \begin{subfigure}[b]{0.32\textwidth}
        \centering
        \includegraphics[width=\textwidth]{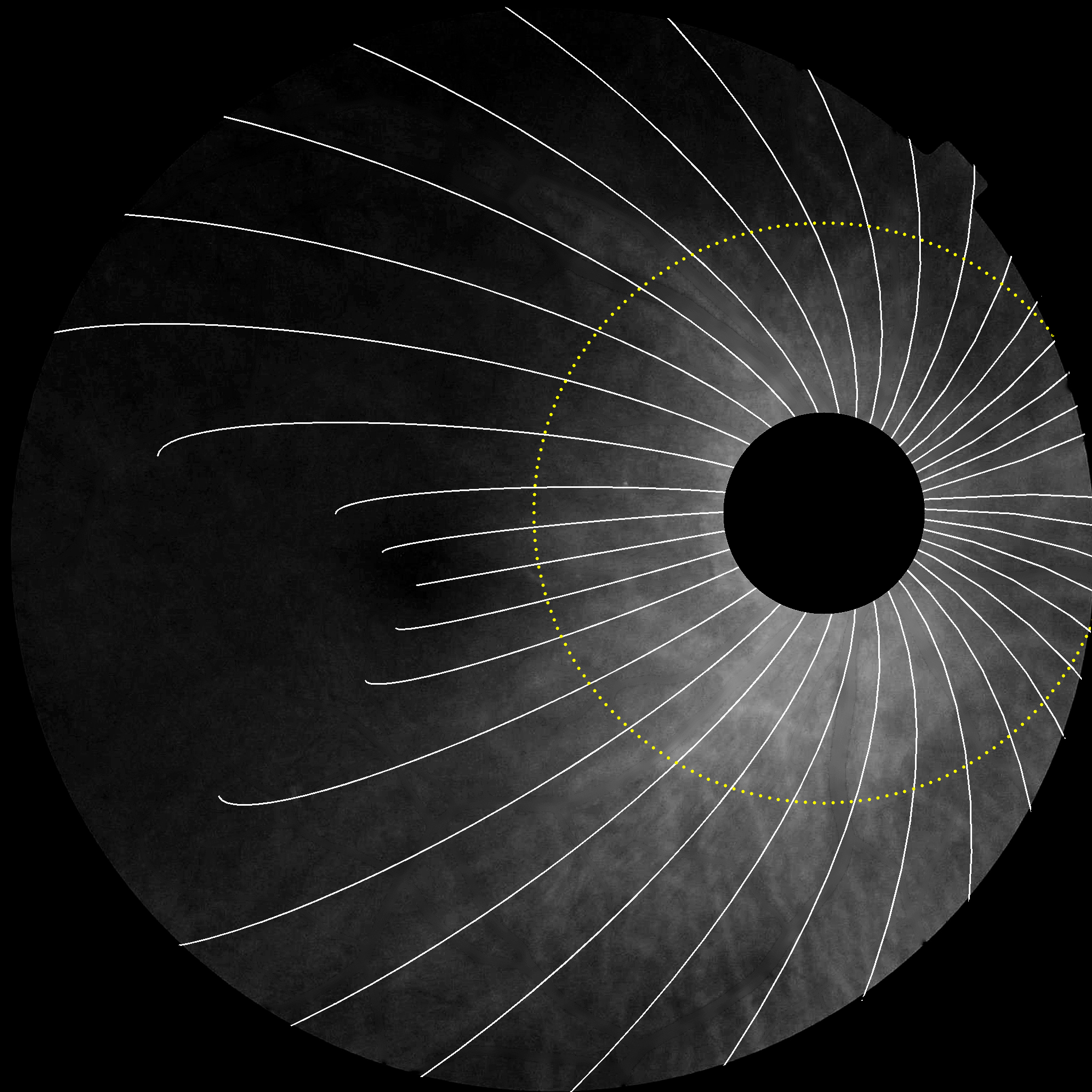} % Replace with your second image file
        \subcaption{}
        \label{fig:image5}
    \end{subfigure}
    \hfill
    % First row, third image
    \begin{subfigure}[b]{0.32\textwidth}
        \centering
        \includegraphics[width=\textwidth]{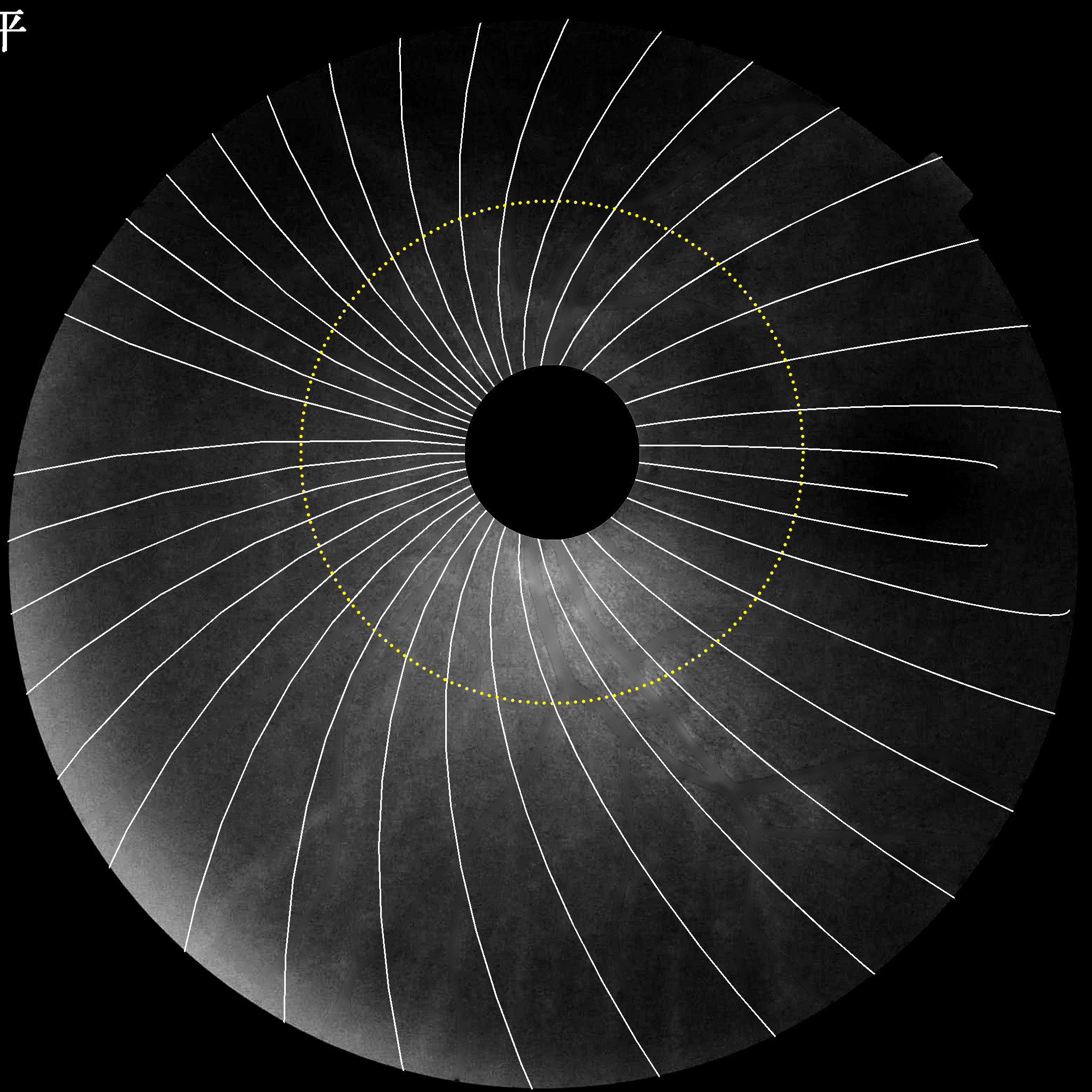} % Replace with your third image file
        \subcaption{}
        \label{fig:image6}
    \end{subfigure}
    
    \vskip\baselineskip
    
    % Second row, first image
    \begin{subfigure}[b]{0.32\textwidth}
        \centering
        \includegraphics[width=\textwidth, height=2cm]{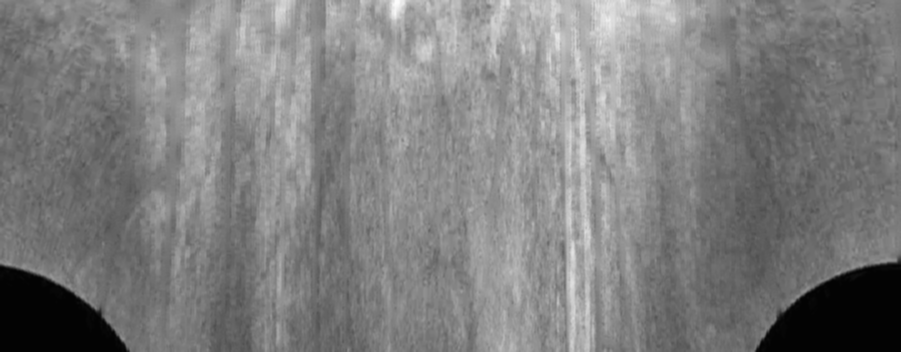} % Replace with your fourth image file
        \subcaption{}
        \label{fig:image7}
    \end{subfigure}
    \hfill
    % Second row, second image
    \begin{subfigure}[b]{0.32\textwidth}
        \centering
        \includegraphics[width=\textwidth, height=2cm]{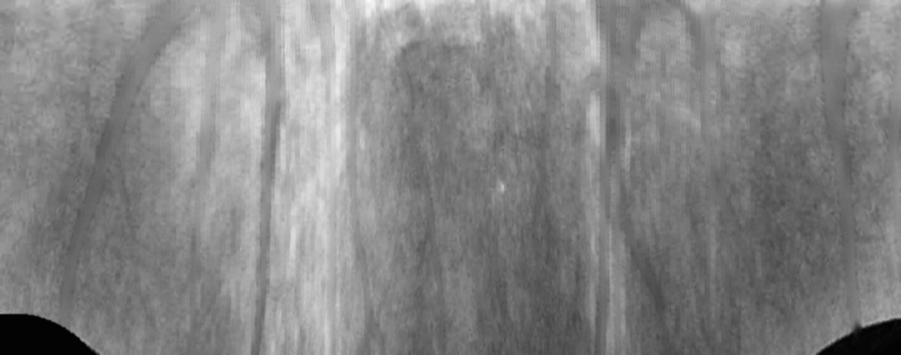} % Replace with your fifth image file
        \subcaption{}
        \label{fig:image8}
    \end{subfigure}
    \hfill
    % Second row, third image
    \begin{subfigure}[b]{0.32\textwidth}
        \centering
        \includegraphics[width=\textwidth, height=2cm]{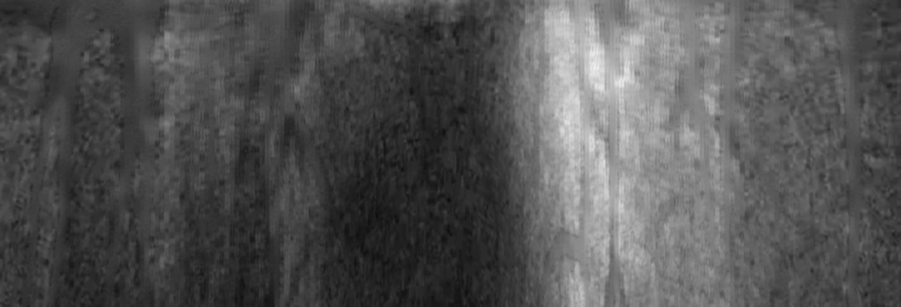} % Replace with your sixth image file
        \subcaption{}
        \label{fig:image9}
    \end{subfigure}

    \caption{
    (a), (b), (c) Illustrations of the RNFL trajectories at $5^\circ$ intervals for various positions of the optic disc and macula. (d), (e), (f) Corresponding results of the ellipse-based polar transformation applied to these trajectories.
    }

    \label{fig:ellipse_based_polar_transform}
\end{figure}

The term $\frac{C_{\text{max}}}{r}$ in Equation \ref{eq:major_axis} encapsulates the phenomenon wherein the major axis length increases as the optic disc center approaches the retina center, thereby broadening the curvature. In the term incorporating a sigmoid function, the inflection point is determined via a binary search algorithm. The inflection point is the point at which the curvature changes significantly—is a critical factor in determining the shape of the ellipse as it fits to the RNFL trajectory. This inflection point corresponds to the transition from the papillomacular bundle to the arcuate bundle, and it is located at a precise angular offset of $30^\circ$ \cite{huh2023papillomacular}. Binary search is used to find the angle corresponding to the inflection point efficiently. Ultimately, the precise determination of the inflection point improves the accuracy of the ellipse model in representing the RNFL structure. The outputs are shown in Figure \ref{fig:ellipse_based_polar_transform}.

\subsection{Deep Learning Inspired Glaucoma Detection}
\label{subsec:deep-learning}
The annular peripapillary region is precisely segmented from the green channel of the fundus image using a mask and crop technique that excludes the optic disc and the area extending beyond 70\% of the distance between the optic disc center and the macula fovea. This process ensures the isolation of the peripapillary zone, which is further refined through blood vessel removal, leaving a texture solely composed of RNFL bundles. By removing the blood vessels, the analysis achieves a focused examination of RNFL integrity, which serves as a critical biomarker for glaucoma diagnosis, independent of the optic disc's structural features.

\begin{figure}[H]
    \centering
    \includegraphics[width=1\textwidth]{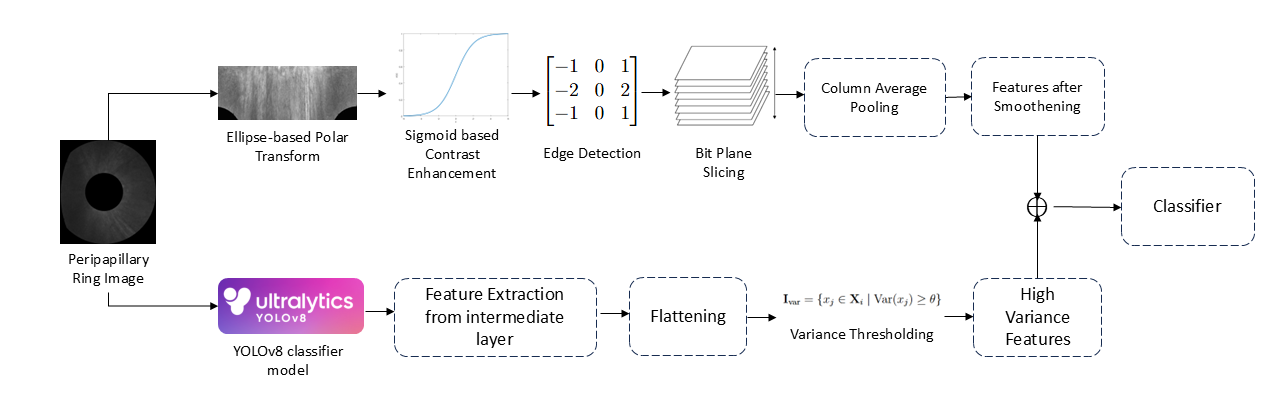} % Replace with your image file
    \caption{Deep Learning Inspired Feature Fusion Pipeline.}
    \label{fig:deep_learning_flow}
\end{figure}

This deep learning-inspired methodology utilizes a fusion strategy, integrating features derived from deep learning models with those extracted through image processing techniques for machine learning applications, resulting in enhanced diagnostic accuracy. Fusion essentially amalgamates the representational power of deep learning with the precision of manually engineered features, ensuring a thorough characterization of the RNFL. This synergy is critical for identifying subtle, glaucoma-specific alterations that may elude detection by either technique independently, thereby optimizing diagnostic sensitivity and specificity in glaucoma detection. Figure \ref{fig:deep_learning_flow} illustrates the flow of this approach.

One branch of the methodology rigorously analyzes the peripapillary region image, with a particular focus on the optic disc’s annular structure, a critical feature in the pathophysiology of glaucoma. This region is often subject to thinning or distortion in glaucomatous degeneration, making it crucial for early diagnostic detection. To model the complex structural and morphological variations associated with glaucomatous changes, we employ the YOLOv8s-cls architecture \cite{ultralytics}, applying transfer learning for model refinement. The model is trained on the set of peripapillary images \( \mathbf{I}_i \), where \( i \) represents the image index, for \( E = 100 \) epochs with a batch size \( b = 16 \) and image resolution \( r = 640 \times 640 \).

The optimization of the model's weight parameters \( \mathbf{W} \) during training results in the extraction of feature vectors \( \mathbf{F}_i \) from the adaptive average pooling layer within the Classify block of the YOLOv8s-cls architecture. The resulting feature vector for each image \( \mathbf{F}_i \in \mathbb{R}^{1280} \), with a dimensionality of 1280, captures significant structural attributes indicative of glaucomatous alterations, particularly those related to the peripapillary ring's morphology. To further refine the feature set, variance thresholding \cite{cuturi2013meanreversion} is applied as expressed mathematically in Equation \ref{eq: variance}.

\begin{equation}
\mathbf{I}_{\text{var}} = \left\{ x_j \in \mathbf{X}_i \mid \text{Var}(x_j) \geq \theta \right\}
\label{eq: variance}
\end{equation}

\begin{equation}
\text{Var}(x_j) = \frac{1}{n} \sum_{i=1}^n (x_{ij} - \bar{x_j})^2
\label{eq:var}
\end{equation}

\( \mathbf{X}_i \) represents the feature vector for the \(i\)-th image. \( x_j \) refers to the \(j\)-th feature in the feature vector \( \mathbf{X}_i \). \( \text{Var}(x_j) \) in Equation \ref{eq:var} denotes the variance of the \(j\)-th feature, which quantifies its variability. \( \theta \) is the variance threshold, which is chosen to be 0.5 to retain the most discriminative features. \(\bar{x_j}\) is the mean of the \(j\)-th feature across all \(n\) images.

The other branch of the methodology focuses on processing the RNFL bundles, where the first step involves straightening the peripapillary structure to \( \mathbf{I}_{\text{polar}} \) using the Ellipse-based Polar Transformation. To enhance glaucomatous patterns in the RNFL bundles, an intensity adjustment is applied to the polar-transformed image \( I_{\text{polar}} \). This transformation aims to accentuate variations around a specific intensity range by compressing lower and higher intensities while emphasizing mid-range differences. A sigmoid function is employed due to its smooth, nonlinear nature and its ability to enhance contrast around a chosen threshold.

The standard sigmoid function is defined in Equation \ref{eq:sig}:
\begin{equation}
    S(x) = \frac{1}{1 + e^{-x}}
\label{eq:sig}
\end{equation}

To adapt the sigmoid for intensity adjustment, we introduce a slope parameter \( \lambda > 0 \) to control the steepness of the transition and a centering term \( \theta \) to shift the sigmoid along the intensity axis as in Equation \ref{eq:sig1}.
\begin{equation}
    S(x) = \frac{1}{1 + e^{-\lambda(x - \theta)}}
\label{eq:sig1}
\end{equation}

We define the input to the sigmoid as the difference between the pixel intensity and the mean intensity of the image, with a tunable bias term \( \delta \) in Equation \ref{eq:sig2}.

\begin{equation}
    x = I_{\text{polar}} - I_{\text{mean}} + \delta
\label{eq:sig2}
\end{equation}

Substituting \( x \) into the Equation \ref{eq:sig1}, the intensity adjusted image \( I_{\text{adjusted}} \) changes to \ref{eq:sig3}.

\begin{equation}
    I_{\text{adjusted}} = \frac{1}{1 + e^{-\lambda (I_{\text{polar}} - I_{\text{mean}} + \delta)}}
\label{eq:sig3}
\end{equation}

This formulation enhances contrast centered around the adjusted mean intensity, where \( \delta \) is a bias parameter that shifts the enhancement focus.

In our implementation, we empirically set \( \delta = -0.5 \), as it yielded optimal enhancement in this dataset for glaucomatous RNFL variations. For other datasets this parameters should need to be adjusted for best analytical outcome. Substituting \( \delta \) into the equation it gives the final Equation \ref{eq:sigmoid_adjustment}.
\begin{equation}
    I_{\text{adjusted}} = \frac{1}{1 + e^{-\lambda (I_{\text{polar}} - I_{\text{mean}} - 0.5)}}
    \label{eq:sigmoid_adjustment}
\end{equation}

where \( \lambda = 0.5 \) controls the steepness of intensity transition. This adjusted image is subsequently used for further analysis, including edge detection and region-based quantification.
 Following intensity adjustment, Sobel edge detection \cite{owotogbe2019edge} is applied to the adjusted image to highlight intensity gradients, crucial for identifying areas where the RNFL bundle exhibits the highest intensity change, particularly in the supratemporal and infratemporal regions, which are critical for glaucomatous analysis. This operation is defined in Equation \ref{eq:edge}.

\begin{equation}
\mathbf{I}_{\text{edges}} = \nabla_{\text{Sobel-X}}(\mathbf{I}_{\text{adjusted}}) = \mathbf{I}_{\text{adjusted}} * \mathbf{K}_x
\label{eq:edge}
\end{equation}

\[
\mathbf{K}_x = \begin{bmatrix}
-1 & 0 & 1 \\
-2 & 0 & 2 \\
-1 & 0 & 1
\end{bmatrix}
\]

where \( \nabla_{\text{Sobel-X}} \) denotes the Sobel operator that computes the gradient of the image and emphasizes the edges of structures in the image.

Subsequently, bit-plane slicing is employed as a crucial technique to extract the most significant bit (MSB) from the image, which preserves the essential structural information necessary for glaucomatous analysis. This step forms the backbone of the entire analysis, as the MSB encapsulates the most vital features that are pivotal for identifying key markers of glaucoma. As highlighted in \cite{10450051}, bit-plane slicing can be mathematically expressed as follows, providing the foundation upon which all further processing and diagnostic conclusions are built.

\begin{equation}
\mathbf{I}_{\text{MSB}} = \text{bit-plane}(\mathbf{I}_{\text{edges}}, \text{MSB}) = \left\lfloor \frac{\mathbf{I}_{\text{edges}}}{2^{n-1}} \right\rfloor \bmod 2
\label{eq:bit_plane}
\end{equation}

In Equation \ref{eq:bit_plane}, the variable n represents the number of bits used to represent the pixel intensity values in the image. After bit-plane slicing, column average pooling is applied to the resulting image \cite{Gholamalinezhad2020PoolingMI}. This operation aggregates the pixel values in each column to produce a single value per column, which helps in capturing the vertical patterns characteristic of RNFL bundles. The pooling is restricted to columns with average of column aggregated pixel values of x-coordinate \( x_1 \) and\( x_2 \) as this range is found in our study crucial for glaucoma detection. The column average pooling operation is shown in Equation \ref{eq:col_avg}.

\begin{equation}
\mathbf{F}_{\text{column avg}} = \frac{1}{M} \sum_{i=1}^{M} \mathbf{I}_{\text{MSB}}(i, j), \quad \text{for } j \in [x_1, x_2]
\label{eq:col_avg}
\end{equation}

where \( \mathbf{F}_{\text{column avg}} \) represents the feature vector obtained from column averaging, \( \mathbf{I}_{\text{MSB}}(i, j) \) is the pixel intensity at row \( i\) column \( j \), and \( M \) is the number of pixels averaged in the column. 

\begin{figure}[H]
    \centering
    \includegraphics[width=1\textwidth]{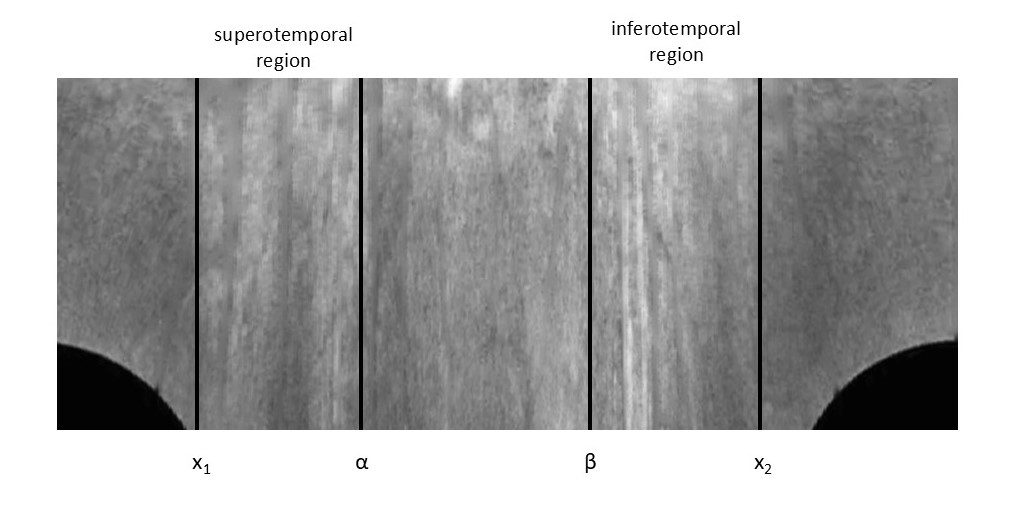} % Replace with your image file
    \caption{Region of Interest in Polar Transformed Fundus Image. \(x_1\) and \(x_2\) mark the lateral bounds; \(\alpha\) and \(\beta\) indicate the centers of the superotemporal and inferotemporal RNFL bundles.}
    \label{fig:region}
\end{figure}

The selected x-coordinate range \([x_1, x_2]\) corresponds anatomically to the region encompassing the superotemporal and inferotemporal RNFL bundles as shown in Figure \ref{fig:region}, particularly after standard preprocessing steps like cropping and alignment around the optic disc. These two arcuate regions are known to exhibit the highest reflectivity and thickness in RNFL due to dense axonal bundles, making them critically important for detecting early glaucomatous damage.  In our study, the cropping procedure was carefully designed to center around the optic nerve head and isolate the vertical band that includes the peak RNFL signal zones, typically lying between these lateral bounds (\(x \approx 250\) to \(650\) pixels) in our standardized imaging setup.

Then, the smoothed gradient is computed using a 1D convolution with a window size of 10 to remove noise and retain the primary gradient features associated with glaucomatous changes. The smoothing operation can be represented in Equation \ref{eq:smooth}.

\begin{equation}
\mathbf{I}_{\text{smoothed}}(j) = \sum_{k=1}^{10} \mathbf{F}_{\text{column avg}}(j+k-5) \cdot \mathbf{K}_{\text{window}}(k)
\label{eq:smooth}
\end{equation}

where \( \mathbf{K}_{\text{window}} \) is a 1D smoothing kernel with a window size of 10,  defined as a normalized box filter. The resulting smoothed feature map \( \mathbf{I}_{\text{smoothed}} \) highlights the primary gradient features while suppressing noise, crucial for identifying glaucomatous changes in the image, as shown in Figure \ref{fig:image_processing} with blue lines.

The entire process, from the ellipse-based polar transformation to the column average pooling and smoothing, yields a set of refined features that are then used for further machine learning analysis to identify glaucomatous alterations with high diagnostic precision.

After extracting the variance-filtered features (\( \mathbf{I}_{\text{var}} \)) and smoothed gradient features (\( \mathbf{I}_{\text{smoothed}} \)), these features are fused to form a unified representation in Equation \ref{eq:fused}.

\begin{equation}
\mathbf{I}_{\text{fused}} = \mathbf{I}_{\text{var}} \oplus \mathbf{I}_{\text{smoothed}}
\label{eq:fused}
\end{equation}

This fusion step integrates both structural and dynamic information, which are essential for capturing glaucomatous changes, particularly in the peripapillary region and RNFL bundles.

The fused feature vector is then normalized to ensure all features are on the same scale as shown in Equation \ref{eq:norm}, facilitating proportional contributions from each feature:

\begin{equation}
\mathbf{I}_{\text{normalized}} = \frac{\mathbf{I}_{\text{fused}} - \mu}{\sigma}
\label{eq:norm}
\end{equation}

Where \( \mu \) and \( \sigma \) represent the mean and standard deviation of the fused vector, respectively. Normalization helps enhance the model's ability to distinguish subtle glaucomatous changes in the optic nerve head and RNFL.

Finally, the normalized feature vector is input into two classifiers, Support Vector Machine (SVM) \cite{Suthaharan2016} and Decision Tree (DT) \cite{song2015decision}, for the final diagnostic prediction. 

\begin{equation}
\hat{y}_{\text{SVM}} = \text{SVM}(\mathbf{I}_{\text{normalized}})
\end{equation}

\begin{equation}
\hat{y}_{\text{DT}} = \text{Decision Tree}(\mathbf{I}_{\text{normalized}})
\end{equation}

The SVM model was configured with a Radial Basis Function (RBF) kernel to enable non-linear decision boundaries capable of capturing complex relationships within the high-dimensional feature space. The regularization parameter \( C \) was set to 100, which imposes a higher penalty on misclassified samples and thus encourages the model to fit the training data more closely. This choice was made to prioritize classification accuracy, particularly given the imbalanced and subtle nature of glaucomatous features. Furthermore, the kernel coefficient \( \gamma \) was set to 0.001, which controls the influence of individual training samples. A lower gamma value leads to a smoother decision boundary, thereby enhancing the model's generalization capability.

The DT classifier was employed using default hyperparameters due to its inherent interpretability and efficiency in handling structured input data. The splitting criterion was set to Gini impurity, which measures the quality of a split based on the homogeneity of the target classes within subsets. The maximum depth of the tree was left unrestricted, allowing the tree to expand fully until all leaves were pure or contained fewer than the minimum number of samples required for a split. This setting ensures that the model can learn complex hierarchical decision rules. The minimum number of samples required to split an internal node was set to 2, and the splitter parameter was set to \texttt{best}, which chooses the optimal feature and threshold that yields the highest information gain at each node. These configurations enable the DT model to adaptively learn from the input features and make rule-based predictions that are both transparent and easy to interpret—particularly useful in clinical decision support systems.

These classifiers are trained to detect glaucomatous pathology, assisting in early-stage detection and enabling timely medical interventions.

\subsection{Image Processing based Glaucoma Detection}
\label{subsec:image-processing}

A rule based image processing feature extraction method is used to enable precise RNFL structural analysis for early glaucoma detection. They are computationally efficient and effective with smaller datasets, making them ideal for resource-limited settings. 

The process described in Algorithm \ref{Image_processing_algo} calculates a dynamic threshold \( T \) based on the smoothed image data \( I_{\text{smoothed}}(x) \) using Equation \ref{eq:dynamic_threshold}.

\begin{equation}
T = \sqrt{\frac{1}{N} \sum_{x=1}^{N} \left( I_{\text{smoothed}}(x) \right)^2} 
+ \lambda \frac{\max(I_{\text{smoothed}}(x)) - \min(I_{\text{smoothed}}(x))}{\frac{1}{N} \sum_{x=1}^{N} I_{\text{smoothed}}(x)}
\label{eq:dynamic_threshold}
\end{equation}

\begin{algorithm}[H]
\caption{Image Processing based Glaucoma Detection}
\label{Image_processing_algo}
\begin{algorithmic}[1]
\STATE \textbf{Input:} Smoothed intensity profile \( I_{\text{smoothed}}(x) \), where \( x \) represents column indices.
\STATE \textbf{Output:} Classification as \textit{Glaucoma} or \textit{Normal}.
\STATE Compute the number of columns \( N \).
\STATE Calculate the threshold:
\[
T = \sqrt{\frac{1}{N} \sum_{x=1}^{N} \left( I_{\text{smoothed}}(x) \right)^2} 
+ \lambda \cdot \frac{\max(I_{\text{smoothed}}(x)) - \min(I_{\text{smoothed}}(x))}{\frac{1}{N} \sum_{x=1}^{N} I_{\text{smoothed}}(x)}.
\]
\STATE Assign \( C(x) = 1 \) if \( I_{\text{smoothed}}(x) \geq T \), else \( C(x) = 0 \).
\STATE Compute the percentage of columns with \( C(x) = 1 \) in the range \( x \in [\alpha, \beta] \), where \( \alpha \) and \( \beta \) denote the approximate boundaries of the temporal region:
\[
P_1 = \frac{\sum_{x=\alpha}^{\beta} C(x)}{\beta - \alpha + 1}.
\]
\STATE If \( P_1 \geq 0.3 \), classify as \textit{Glaucoma}. 
\STATE Else, find \( x_{\text{min}} = \arg \min_{x} I_{\text{smoothed}}(x) \).
\STATE \( \text{Left}_1 = \sum_{x=1}^{x_{\text{min}}-1} C(x), \quad 
\text{Right}_1 = \sum_{x=x_{\text{min}}+1}^{N} C(x), \quad
DI = \frac{\text{Left}_1}{\text{Right}_1 + \epsilon} \).
\STATE \text{Classify as \textit{Normal} if \( 0.3 < DI < 3.3 \), else classify as \textit{Glaucoma}}.

\end{algorithmic}
\end{algorithm}

\( N \) represents the total number of columns in the image and \( \lambda = 4 \) is a scaling factor that adjusts the influence of the range-to-mean ratio. In Equation \ref{eq:dynamic_threshold}, the first term is the RMS of the energy term $I_{\text{smoothed}}(x)$ represents the overall intensity variation in the image, aiding in the identification of prominent features that may indicate structural abnormalities. The second term, the range-to-mean ratio, quantifies the contrast within the image, emphasizing areas with significant variation that could signal important changes. The scaling factor \( \lambda\) fine-tunes the sensitivity of the threshold calculation, helping to balance the detection of meaningful variations with the exclusion of irrelevant details.

\begin{figure}[H]
    \centering
    % First image
    \begin{subfigure}[b]{0.45\textwidth}
        \centering
        \includegraphics[width=\textwidth]{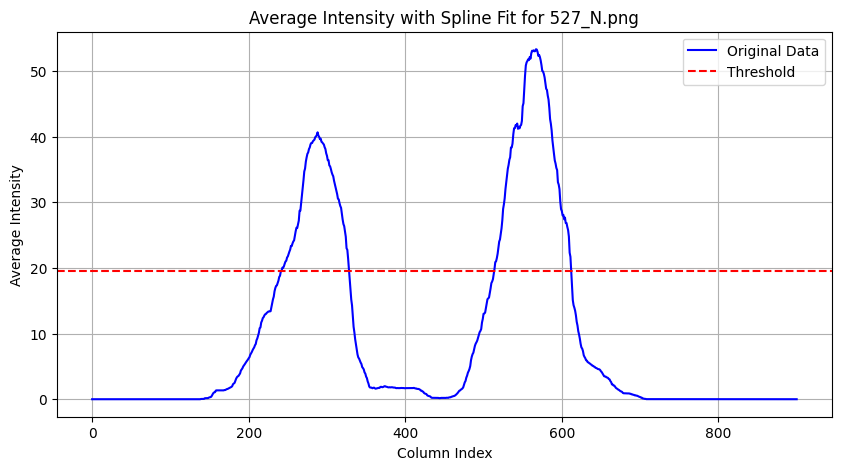} % Replace with your first image file
        \subcaption{}
        \label{fig:image11}
    \end{subfigure}
    \hfill
    % Second image
    \begin{subfigure}[b]{0.45\textwidth}
        \centering
        \includegraphics[width=\textwidth]{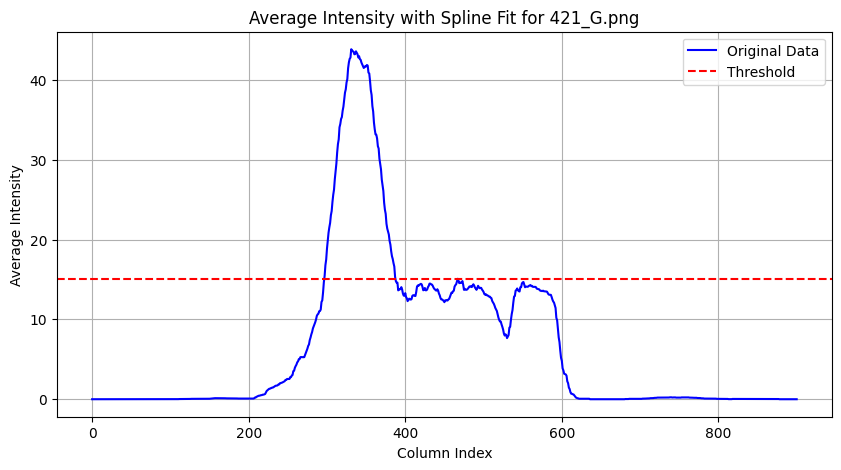} % Replace with your second image file
        \subcaption{}
        \label{fig:image12}
    \end{subfigure}
    
    \caption{
        (a) Column average curve and threshold calculation on normal image. (b) Column average curve and threshold calculation on glaucoma image.
    }
    \label{fig:image_processing}
\end{figure}

The Disparity Index (DI) is crucial for detecting asymmetry in the RNFL. In glaucoma, RNFL thinning often spares the temporal region, leading to masked pixels (\( C(x) = 1 \)) in this area. If more than 30\% of these pixels are masked, it is classified as glaucoma. On the other hand, for diffuse thinning in the supratemporal or infratemporal regions, one side’s mask becomes thicker than the other, creating an imbalance. The DI quantifies this asymmetry, as shown in Figure \ref{fig:image_processing} and if its value deviates from the range \( 0.3 \leq \text{DI} \leq 3.33 \), it is considered as glaucoma. From summary statistics of derived features in Table \ref{tab:sensitivity_di} as Disparity Index (DI) threshold, precision, recall, F1-score, accuracy, and using cross-column insights the threshold of DI is derived as shown in step 10 of Algorithm \ref{Image_processing_algo}.

While the image processing technique demonstrates strong potential in detecting glaucoma, it is important to acknowledge a few areas for improvement. The accuracy of RNFL analysis can be influenced by image quality, with variations in resolution or noise potentially affecting results. Despite these factors, the method holds promise, and further optimization and testing on diverse datasets will enhance its robustness, particularly in detecting early-stage glaucoma. 

\section{Results and Discussions}
\label{sec:results}

As detailed in Section \ref{subsec:preprocessing}, The YOLOv8 model exhibited high performance in detecting the optic disc and macula, as summarized in Table~\ref{tab:results}. 

\begin{table}[H]
\centering
\caption{Performance metrics for optic disc and macula detection.}
\label{tab:results}
\begin{tabular}{|l|c|c|c|c|}
\hline
\textbf{Structure}      & \textbf{Precision} & \textbf{Recall} & \textbf{mAP@50} & \textbf{mAP@50:95} \\ \hline
\textbf{Optic Disc}     & 0.998              & 1.000           & 0.995           & 0.773              \\ \hline
\textbf{Macula}         & 0.978              & 0.981           & 0.987           & 0.572              \\ \hline
\end{tabular}
\end{table}

The U-Net model, trained for vessel segmentation, achieves a training accuracy of 99.92\%, a validation accuracy of 98.02\% and a Dice coefficient of 0.89, confirming its effectiveness as a preprocessing step for RNFL-based glaucoma detection.

\begin{figure}[H]
    \centering
    % First image
    \begin{subfigure}[b]{0.32\textwidth}
        \centering
        \includegraphics[width=\textwidth, height=4cm]{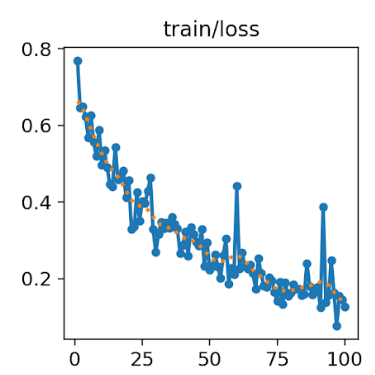} % Replace with your first image file
        \subcaption{}
        \label{fig:image13}
    \end{subfigure}
    \hfill
    % Second image
    \begin{subfigure}[b]{0.32\textwidth}
        \centering
        \includegraphics[width=\textwidth, height=4cm]{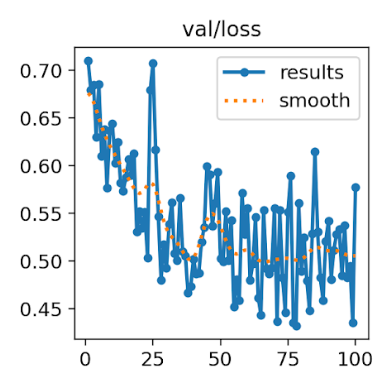} % Replace with your second image file
        \subcaption{}
        \label{fig:image14}
    \end{subfigure}
    \hfill
    % Third image
    \begin{subfigure}[b]{0.32\textwidth}
        \centering
        \includegraphics[width=\textwidth, height=3.9cm]{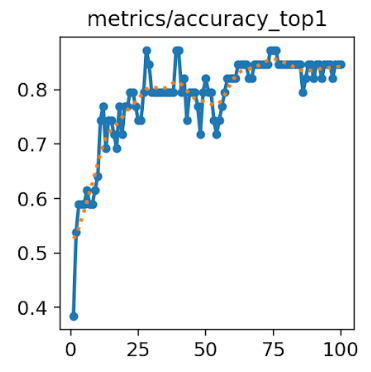} % Replace with your third image file
        \subcaption{}
        \label{fig:image15}
    \end{subfigure}
    
    \caption{
        (a) Training loss per epoch. (b) Validation loss per epoch. (c) Accuracy per epoch.
    }
    \label{fig:loss_accuracy}
\end{figure}

Prior to the classification analysis as detailed in Section \ref{subsec:deep-learning}, a transfer learning approach using YOLOv8 was applied to a dataset of peripapillary ring images for glaucoma detection. The model was trained to extract features relevant to glaucomatous changes, achieving a top-1 accuracy as 87.1\%. The training loss, validation loss and accuracy curves are shown in Figure \ref{fig:loss_accuracy}.

The training performance comparison using the ellipse-based polar transformation (polar), the ring-based features (ring), and the fusion of features is presented in Table \ref{tab:fusion_train_results}, followed by the testing performance in Table \ref{tab:fusion_results}.

\begin{table}[H]
\centering
\caption{Training performance comparison of classifiers using polar, ring, and fusion features.}
\label{tab:fusion_train_results}
\resizebox{\textwidth}{!}{%
\begin{tabular}{|l|l|c|c|c|c|c|}
\hline
\textbf{Feature Set} & \textbf{Classifier} & \textbf{Mean CV} & \textbf{Precision} & \textbf{Recall} & \textbf{F1-Score} & \textbf{Accuracy} \\ \hline
\multirow{2}{*}{\textbf{Ring}} 
    & SVM           & 0.985            & 0.987              & 0.986           & 0.986             & 0.987             \\ \cline{2-7} 
    & Decision Tree & 0.975            & 0.976              & 0.977           & 0.976             & 0.977             \\ \hline
\multirow{2}{*}{\textbf{Polar}} 
    & SVM           & 0.902            & 0.891              & 0.893           & 0.892             & 0.894             \\ \cline{2-7}
    & Decision Tree & 0.854            & 0.842              & 0.837           & 0.839             & 0.838             \\ \hline
\multirow{2}{*}{\textbf{Fusion}} 
    & SVM           & 0.991            & 0.996              & 0.996           & 0.996             & 0.996             \\ \cline{2-7}
    & Decision Tree & 0.980            & 0.982              & 0.983           & 0.982             & 0.983             \\ \hline
\end{tabular}%
}
\end{table}

\begin{table}[H]
\centering
\caption{Performance comparison of classifiers using polar, ring, and fusion features.}
\label{tab:fusion_results}
\resizebox{\textwidth}{!}{%
\begin{tabular}{|l|l|c|c|c|c|c|}
\hline
\textbf{Feature Set} & \textbf{Classifier} & \textbf{Mean CV} & \textbf{Precision} & \textbf{Recall} & \textbf{F1-Score} & \textbf{Accuracy} \\ \hline
\multirow{2}{*}{\textbf{Ring}} & SVM                 & 0.981            & 0.979              & 0.978           & 0.978             & 0.978             \\ \cline{2-7} 
                              & Decision Tree       & 0.960            & 0.964              & 0.964           & 0.964             & 0.965             \\ \hline
\multirow{2}{*}{\textbf{Polar}} & SVM                 & 0.893            & 0.862              & 0.865           & 0.863             & 0.865             \\ \cline{2-7}
                              & Decision Tree       & 0.832            & 0.816              & 0.801           & 0.806             & 0.801             \\ \hline
\multirow{2}{*}{\textbf{Fusion}}& SVM                 & 0.984            & 0.993              & 0.993           & 0.993             & 0.993             \\ \cline{2-7}
                              & Decision Tree       & 0.969            & 0.979              & 0.979           & 0.979             & 0.979             \\ \hline
\end{tabular}%
}
\end{table}

The fusion approach yielded the highest performance, with the SVM achieving a mean CV of 0.984, precision, recall, and F1-score of 0.993, and accuracy of 0.993. The Decision Tree classifier attained a mean CV of 0.969, precision, recall, and F1-score of 0.979, and accuracy of 0.979.

This methodology amalgamates deep learning-derived features with manually extracted features from ellipse-based polar-transformed images. The column average pooling step facilitated the straightening of images, thereby enabling the effective extraction of textural alterations in the peripapillary region, which is pivotal for manual feature delineation. Additionally, this alignment allowed for precise bit plane slicing, further enhancing feature extraction. This meticulous alignment also augmented deep learning features by refining the accuracy of ring-like image feature extraction, thereby bolstering the reliability of glaucomatous change detection. The confluence of these synergistic techniques culminates in a more robust diagnostic tool for glaucoma.

\begin{figure}[H]
    \centering
    \includegraphics[width=0.5\textwidth]{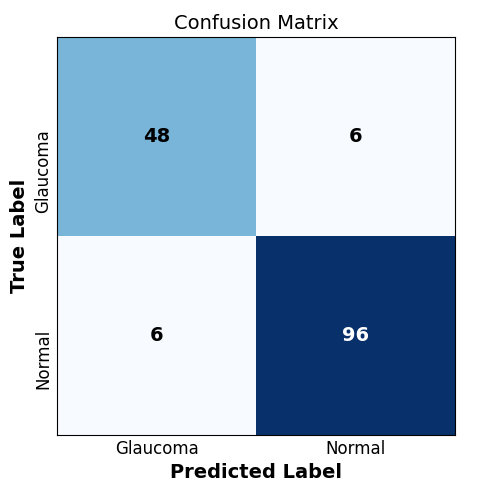} % Replace with your image file
    \caption{Confusion matrix showing TP, FP, FN, TN values for image processing based approach using a set of sample dataset.}
    \label{fig:image_processing_plot}
\end{figure}

In contrast, while this method demonstrates remarkable accuracy, it is computationally intensive, and demands significant time and resources, which may hinder its scalability and widespread applicability. To mitigate these challenges, we have devised an image-processing-based approach, elaborated in Section \ref{subsec:image-processing}, which achieves a harmonious balance between computational efficiency and diagnostic accuracy, attaining a performance of 92.31\%. The corresponding confusion matrix for this approach is presented in Figure \ref{fig:image_processing_plot}.

To determine an appropriate threshold range for the Disparity Index (DI), a statistical analysis was conducted. The results of this statistical analysis are summarized in Table~\ref{tab:sensitivity_di}. As shown, the performance peaks at the range \(0.30 \leq \text{DI} \leq 3.33\), with a notable drop in F1-score and accuracy for both narrower and wider thresholds. This supports our choice of bounds as being optimal for detecting glaucomatous asymmetry without introducing false positives from benign variations.

\begin{table}[H]
\centering
\caption{Summary statistics in DI threshold determination for glaucoma detection performance.}
\label{tab:sensitivity_di}
\begin{tabular}{|c|c|c|c|c|}
\hline
\textbf{Threshold (DI)} & \textbf{Precision} & \textbf{Recall} & \textbf{F1-Score} & \textbf{Accuracy} \\
\hline
0.20–5.00 & 0.7780 & 0.8350 & 0.8055 & 0.8410 \\
0.25–4.00 & 0.8100 & 0.7520 & 0.7799 & 0.8510 \\
\textbf{0.30–3.33} & \textbf{0.8890} & \textbf{0.8890} & \textbf{0.8890} & \textbf{0.9231} \\
0.35–2.86 & 0.7835 & 0.7280 & 0.7546 & 0.8395 \\
0.40–2.50 & 0.7950 & 0.6950 & 0.7412 & 0.8180 \\
\hline
\end{tabular}
\end{table}

The image processing approach significantly reduces per-image execution time compared to deep learning models. While deep learning methods require extensive computational resources and processing time due to the complexity of the networks, the image processing approach operates with considerably lower resource requirements, ensuring faster results. This makes it more suitable for applications where real-time analysis is critical.

In Table \ref{tab:glaucoma_comparison}, this paper's results are compared with other studies to highlight its relative performance. Such comparisons are crucial for evaluating strengths, limitations, and trade-offs between accuracy, efficiency, and practicality, guiding the development of methods suited for specific needs like real-time or resource-constrained applications.

\begin{table}[h]
\centering
\caption{Comparison of recent glaucoma detection methods and their accuracies}
\resizebox{\textwidth}{!}{%
\begin{tabular}{|l|l|c|}
\hline
\textbf{Paper Name} & \textbf{Method} & \textbf{Accuracy} \\ \hline
Kolar and Jan \cite{kolar2008detection} & Fractal Dimensions & 93.8\% \\ \hline
Acharya et al. \cite{acharya2011automated} & Texture and Higher Order Spectra Features & 91\% \\ \hline
Oh et al. \cite{Oh2015AutomaticCD} & Optical Coherence Tomography Angiogram Images & 94.3\% \\ \hline
Prananda et al. \cite{Prananda2023} & DenseNet Features & 92.88\% \\ \hline
Our Study & Image Processing based Approach & 92.31\% \\ \hline
Our Study & Deep Learning inspired Approach & 99.3\% \\ \hline
\end{tabular}
\label{tab:glaucoma_comparison}
}
\end{table}

The comparison in Table \ref{tab:glaucoma_comparison} shows that the image processing approach, with 92.31\% accuracy, provides results comparable to other methods, while the deep learning approach, achieving 99.3\% accuracy, surpasses all other techniques in performance.

\section{Conclusion}
\label{sec:conclusion}

In conclusion, this study demonstrates the adaptive capabilities of our ellipse-based algorithm in detecting glaucomatous changes through its ability to effectively capture variations in the RNFL. The method’s flexibility is evident in the two pipelines it supports. The first, which integrates deep learning-derived features with those manually extracted from the ellipse-based polar-transformed images, achieves an accuracy of 99.3\% with an SVM classifier, providing high diagnostic precision at the cost of increased computational demand. The second pipeline, relying solely on image processing, achieves a competitive accuracy of 92.31\%, while offering a more efficient solution in terms of computational resources, making it suitable for resource-constrained environments.

By offering these two complementary options, our approach strikes a balance between high accuracy and computational efficiency, ensuring adaptability to different clinical needs. These findings validate the utility of our adaptive ellipse-based algorithm and provide a solid foundation for future improvements and real-world applications in glaucoma detection.

While the high accuracy reported is based on a specific dataset, further testing on independent and diverse datasets is necessary to assess generalizability. Future research should focus on validating the model across different datasets and imaging modalities to improve its robustness. Integrating RNFL-based analysis with additional biomarkers, such as the cup-to-disc ratio or SNIT, could strengthen the diagnostic framework.

In conclusion, this study reinforces the importance of RNFL in automated glaucoma detection with CFI images using adaptive transformation and establishes a foundation for developing efficient, clinically relevant diagnostic tools that enhance early detection and patient care.

% \section{Competing Interests}
% \label{sec:Competing Interests}

% Not Applicable

% \section{Funding Information}
% \label{sec:Funding Information}

% Not Applicable

\section{Author contribution}
\label{sec:Author contribution}

Paul S. designed and developed the Ellipse-based Polar Transform algorithm and the deep learning-based glaucoma detection algorithm. Mallick S. developed the image processing-based glaucoma detection algorithm. Sen A. critically reviewed all methodological processes and contributed to the writing and revision of the manuscript.

\section{Data Availability Statement}
\label{sec:Data Availability Statement}

This study utilized the publicly available FIVES dataset. The dataset can be accessed at https://www.nature.com/articles/s41597-022-01564-3.

% \section{Research Involving Human and /or Animals}
% \label{sec:Research Involving Human and /or Animals}

% This research did not involve any direct interaction with human participants or animals. The study was conducted using the FIVES dataset, which is publicly available and anonymized. All data used comply with ethical standards and data protection regulations as outlined by the dataset providers.

% \section{Informed Consent}
% \label{sec:Informed Consent}

% Not Applicable

 \bibliographystyle{elsarticle-num} 
 \bibliography{cas-refs}





% \end{thebibliography}
\end{document}